\documentclass[letterpaper, 10 pt, conference]{ieeeconf}  
\usepackage{xcolor}

\IEEEoverridecommandlockouts                              

\usepackage{graphicx} 
\usepackage{mathptmx} 
\usepackage{amsmath} 
\usepackage{amssymb}  
\usepackage{cite}
\usepackage{stfloats}
\usepackage[linesnumbered,ruled,vlined]{algorithm2e}
\usepackage{eucal}
\usepackage{esvect}
\usepackage{booktabs}
\usepackage{multirow}
\usepackage{hyperref}
\usepackage{comment}
\usepackage{threeparttable} 

\title{\LARGE \bf
RoboHitch: Learning Visual Affordance from Disordered Keypoints for Hitch Knots Tying
}

\author{Jiahui Zuo$^{1}$, Boyang Zhang$^{1}$, and Fumin Zhang$^{1}$, \textit{Fellow, IEEE} 
\thanks{*The work described in this paper was partially supported by grants AoE/E-601/24-N, 16203223, and C6029-23G from the Research Grants Council of the Hong Kong Special Administrative Region, China.}
\thanks{$^{1}$J. Zuo, B. Zhang, and F. Zhang (corresponding author) are with the Department of Electronic and Computer Engineering, The Hong Kong University of Science and Technology, Hong Kong (email:  jzuoai@connect.ust.hk, bzhangcd@connect.ust.hk, eefumin@ust.hk).}%
}
\begin{document}  
\setlength{\textfloatsep}{5pt}

\maketitle

\begin{abstract}
Robotic manipulation of deformable linear objects (DLOs) presents significant challenges due to complex dynamics and frequent self-occlusions. 
Existing robotic knot tying methods typically rely on precise topological state tracking with ordered keypoints and explicit edge connectivity. This reliance makes them prone to failures due to tracking drift and topology mismatch caused by repeated bending and crossings during knot formation.
To address these limitations, we introduce RoboHitch, a novel framework that learns to perform hitch knot tying from human demonstrations using only disordered 3D keypoints and RGB images. This eliminates the need for explicit topological order, allowing for more flexible manipulation. Our method employs a dynamic Graph Autoencoder to extract geometric features from untracked keypoints, complemented by a Convolutional Autoencoder that captures essential visual context. A bidirectional cross-attention mechanism then fuses these modalities to jointly predict pick and place affordances, facilitating implicit reasoning about the rope's state and enabling knot tying under occlusion.
Real-world experiments demonstrate the effectiveness and generalizability of our approach, successfully completing hitch knots in scenarios with self-occlusions.

\end{abstract}


\section{Introduction}
Knotting manipulation of DLOs is widely employed in diverse applications, including surgery, industrial automation, and maritime operations.
Knot tying can be classified into three distinct categories: (1) knots, formed within a single rope like overhand knots, (2) bends, connecting between two or more ropes, and (3) hitches, tying a rope to another object.
In real-world scenarios, knot-tying is typically performed to fasten a rope to a target object rather than creating an isolated knot. Consequently, this work focuses on robotic hitch knot tying, a task that humans perform with ease, but remains difficult for robotics.

A fundamental challenge in robotic knot tying is DLO state representation.
Unlike rigid objects, there is no direct representation of the DLO state\cite{doi:10.1126/scirobotics.adt1497}.
Although RGB images and point clouds are easy to obtain, they are relatively unstructured and often data hungry. 
To facilitate subsequent motion planning, DLOs are typically represented as ordered lists of keypoints, where the order encodes topology along the rope. 
This structured prior is useful, but accurate keypoint estimation is difficult during knot formation because self-occlusion can break correspondence and ordering.
While previous work has made progress in tracking DLO topology using point cloud registration for overhand knots or simpler configurations\cite{myronenko2010point, tang2022track, xiang2023trackdlo}, these methods often assume uncrossed DLO initialization and struggle to maintain accurate keypoint tracking from complex self-occlusion states for knot tying\cite{caporali2022fastdlo, keipour2022deformable}. Topology mismatch from tracking failure can easily lead to task failure caused by failed grasps.
With a DLO state estimate, both model-based\cite{tang2022track, dinkel2025knotdlo, tang2018framework} and learning-based methods\cite{suzuki2021air, combing2017, peng2024tiebot} seek the optimal action sequences to complete the task. 
Model-based pipelines typically require explicit and consistent keypoint ordering (often via markers) to classify knot states\cite{manturov2018knot} and execute hand-designed primitives to transition states. The reliance on explicit topological states making them sensitive to occlusion-induced tracking errors without markers. 
\begin{figure}[t]  
    \centering
    \includegraphics[width=1.0\linewidth]{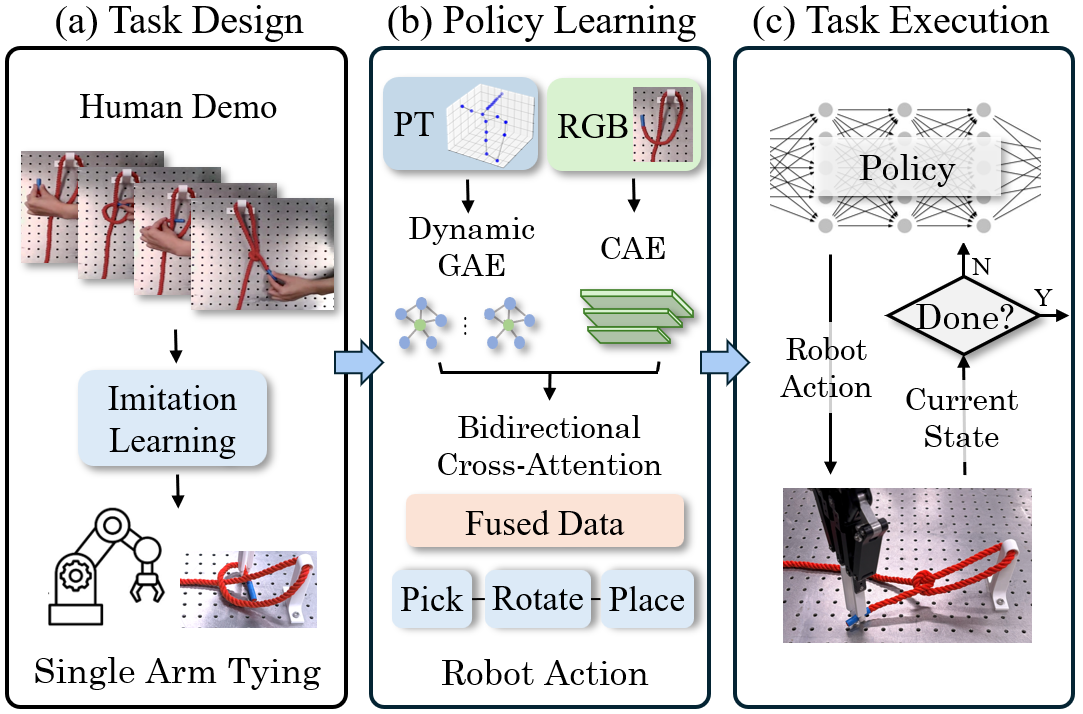}
    \caption{Overview of our study. (a) Our approach learns to tie hitch knots through human demonstrations. (b) Given an RGB image and 3D keypoints of the rope, the model employs bidirectional cross-attention to fuse multimodal features and predict pick, rotate, and place affordances. (c) The robot executes sequential actions until the task is completed.    
    }
    \label{first}
\end{figure}
Furthermore, many learning-based methods use RGB or point cloud inputs from easier state estimation and adopt end-to-end frameworks that tightly couple perception and action. 
This coupling often leads to large quantities of real-world data requirements and makes policies difficult to interpret.

As shown in Fig.~\ref{first}, we propose a framework that learns hitch knot tying policies from visual observations, without relying on perfectly tracked or ordered topological states. Our key insight is that by fusing disordered 3D rope keypoints and RGB scene image, the model can implicitly reason about the rope's physical state and infer effective manipulation actions with limited data, even in the presence of occlusions or crossings. This multimodal bias combines geometric and visual cues, thus constraining the complexity of learning and facilitating more efficient model training. 
Specifically, we employ a Graph Autoencoder (GAE) to encode the topological features from disordered keypoints, while a Convolutional Autoencoder (CAE) processes scene features derived from RGB input. A bidirectional cross-attention mechanism further integrates both modalities to jointly predict pick and place affordances tailored for single-arm robotic knot tying.
We collect human demonstration videos to train our policy and evaluate its performance through a series of real-world hitch knot-tying experiments. The results show that our method achieves robust knotting performance across various scenarios, effectively handling challenges related to disordered rope point perception and self-occlusions.

Key contributions of this work include the following:
\begin{itemize}
\item 
We present a rope state representation that extracts disordered 3D keypoints from a potentially occluded rope, suitable for downstream manipulation without fragile tracking of precise topological order.
\item 
We introduce a knotting framework that leverages bidirectional cross-attention to fuse rope geometric features and visual scene features, facilitating the prediction of pick and place affordances for hitch knot tying in visually complex environments.
\item 
The framework is experimentally validated through real-world robotic hitch knot tying with a single arm, demonstrating effective performance under self-occlusion.

\end{itemize}


\section{Related Work}

\subsection{State Detection and Tracking of DLOs}
Accurate DLO topology estimation for knot tying remains challenging, since vision systems often fail to maintain ordered correspondence under self-occlusion and complex self-crossings.
Some works simplify the problem by assuming the rope is fully visible to track during manipulation \cite{9341330, 9812244, 9888782}.
Significant progress has been made in developing visual tracking methods that can handle some occlusions and self-crossings\cite{9721686, 9561012, xiang2023trackdlo}.
A common pipeline first extracts an uncrossed initial state via segmentation and refinement \cite{caporali2022fastdlo, keipour2022deformable}, subsequently tracks the consistent rope points across frames using non-rigid registration techniques such as Coherent Point Drift (CPD)\cite{myronenko2010point, tang2022track, xiang2023trackdlo} or predictive models \cite{dinkel2025dlo}.
However, during knot formation, persistent bending and repeated crossings can cause drift and topology mismatch, leading to incorrect state estimates and downstream manipulation failures \cite{xiang2023trackdlo, dinkel2025dlo}.
Unlike these tracking approaches, our method foregoes the need for a perfectly ordered topological sequence. Instead, we demonstrate that a robust knot-tying policy can be learned directly from a disordered set of keypoints inputs, which is more reliable in self-occluded scenarios.




\subsection{Robotic Knotting of Ropes}
Robotic knot-tying research can be broadly categorized into two main approaches, model-based methods that utilize estimated geometric information \cite{tang2018framework, dinkel2025knotdlo, 1242193} and learning-based approaches that employ reinforcement or imitation learning \cite{suzuki2021air, combing2017, peng2024tiebot}.
Model-based methods typically rely on physically accurate simulations or geometric reasoning based on the estimated state of the rope. Some approaches plan actions by computing the similarity between the current state and a predefined goal state using registration techniques like CPD \cite{tang2018framework}. Recent study has also explored the application of Reidemeister Moves principles from knot theory\cite{manturov2018knot} to define explainable action primitives for state transitions \cite{dinkel2025knotdlo}. A key drawback of these methods is their heavy reliance on precisely tracked state estimation, which is difficult to guarantee in knotting.
On the other hand, learning-based methods develop knot-tying policies in a data-driven manner \cite{combing2017, yamakawa2013dynamic, suzuki2021air}. These strategies can reduce the dependency on precise models and explicit state representations. However, many previous learning approaches are end-to-end, mapping pixel inputs directly to actions, which often results in limited interpretability and a high demand for training data.
Our work bridges these two paradigms. We employ a learning-based framework trained on human demonstrations and structure it around an interpretable multimodal state representation that incorporates both disordered rope keypoints and RGB images. This allows the policy to learn actions from human examples effectively with limited data while retaining greater explainability than purely end-to-end methods.

Furthermore, studies on robotic knot-tying can be distinguished by their operational environments: some focus on knotting on a workbench, where tabletop constraints simplify the task, while others address in-air manipulation, which demands more sophisticated handling of unconstrained rope dynamics \cite{kudoh2015air, seo2019study, suzuki2021air}.
In our hitch-tying scenario, the rope is partially supported on a table and partially suspended in air, looped through a hole in a pole, requiring a policy that can reason about both supported and unsupported sections.

\section{Problem Formulation}
We formulate the task of single-arm hitch knot tying as a sequential decision-making problem with partial observability. 
The policy $\pi$, trained on human demonstration videos of state-action pairs ${(s_t, a_t)}$, aims to encode the implicit expert dynamic patterns and predict the action
\begin{equation}
a_t = \pi(s_t)
\end{equation}
which drives the rope toward the desired knotted state. 
Each action $a_t$ is parameterized as a 3-tuple:
$(i_t, \theta_t, p_t)$, where $i_t \in \{1, \dots, N_k\}$ 
denotes the index of the selected rope keypoint for grasping, $\theta_t \in [-\pi/2, \pi/2]$ specifies the in-plane rotation for gripper orientation during placement, and $p_t \in \mathbb{R}^2$ represents the pixel coordinates for the target placement location. 

\begin{figure}[t]
\centering
\includegraphics[width=1.0\linewidth]{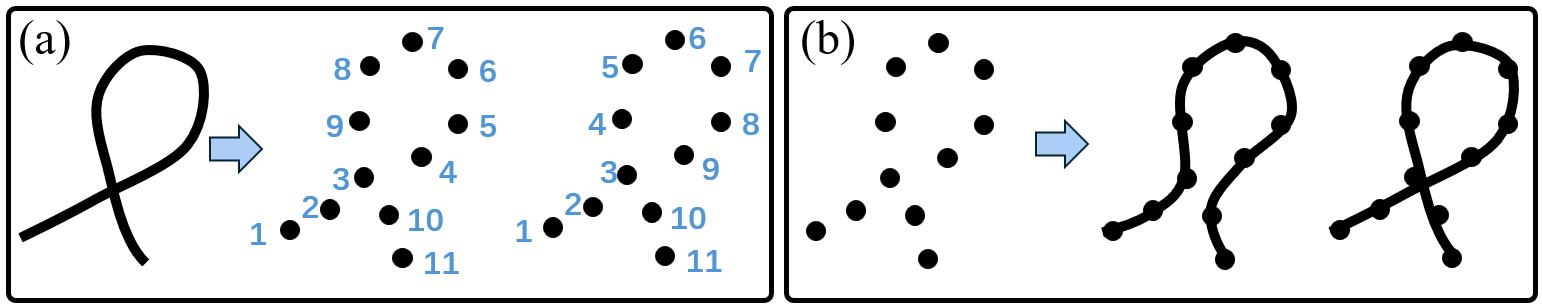}
\caption{Challenges in clustering-based rope state estimation without order tracking. (a) \textit{Point Set Permutation}: Identical states can result in different keypoint sequences. (b) \textit{Topological Ambiguity}: Multiple rope configurations may correspond to the same keypoint set.}
\label{prob}
\end{figure}

Since we do not track the fragile topological sequence order of the rope, two major challenges arise: 
\begin{itemize}
\item 
Point Set Permutation problem: Due to the lack of consistent correspondence across observations, identical rope states may be represented by different keypoint sequences, as shown in Fig.~\ref{prob}(a). This inconsistency misleads models that rely on fixed keypoint ordering or predefined edge connectivity. 
\item 
Topological Ambiguity problem: Resulting from the discrete keypoint representation, distinct rope configurations can map to identical sets of points, as shown in Fig.~\ref{prob}(b). This ambiguity requires the model to distinguish the underlying topology from disordered points without explicit connectivity.
\end{itemize}

The first challenge is addressed through our permutation-invariant GAE while the second one is solved with the multimodal fusion mechanism introduced in Section \ref{sec3_3}.
Specifically, we represent the rope state using a dynamic graph structure. The multimodal state input of the policy is defined as
\begin{equation}
s_t = f(x^\text{pt}, x^\text{rgb})
\end{equation}
where $x^\text{pt} \in \mathbb{R}^{N_ k \times 3} $ denotes the untracked and disordered rope keypoints, and $x^\text{rgb} \in \mathbb{R}^{H \times W \times 3}$ represents the corresponding RGB image that provides rich contextual information.
By integrating both geometric and visual information, the multimodal state $s_t$ helps overcome perceptual limitations and supports the effective prediction of the action $a_t$.


\section{Rope State Estimation}
\label{sec3_2}
The rope state estimation module extracts the rope keypoints $x^\text{pt}$ and their corresponding grasping orientations from the rope point cloud data collected by an RGB-D camera. 
Accurate rope keypoint registration across frames is challenging due to frequent self-occlusions during the knotting process. Thereby we extract a disordered set of rope keypoints $x^\text{pt}$ through a clustering approach, avoiding reliance on error-prone tracking of an ordered sequence.
The rope point cloud $\mathcal{P} = \{p_j\}_{j=1}^{N}$ is initially segmented into $M$ spatially coherent clusters $\{\mathcal{S}_m\}_{m=1}^M$ using Density-Based Spatial Clustering of Applications with Noise (DBSCAN) \cite{schubert2017dbscan},  an unsupervised clustering algorithm which can discover clusters of arbitrary shape and density. For each cluster $\mathcal{S}_m$, we employ K-means clustering to generate $K_m$ representative centroids $\mathcal{C}_m = \{c_{m,k}\}_{k=1}^{K_m}$. The union of these centroids forms our candidate keypoint set, defined as $x^\text{pt} = \bigcup_{m=1}^M \mathcal{C}_m$.

To determine the local grasping orientation at each keypoint, we utilize Singular Value Decomposition (SVD) rather than predicting it directly, which reduces the model's complexity. For a human-selected (or model-predicted) point $p_{pick}$, we first identify its nearest neighbor $p_c$ within the point cloud $\mathcal{P}$.
A local neighborhood $\mathcal{N}$ is then formed by extracting all points within a radius $r$ of $p_c$
\begin{equation}
    \mathcal{N} = \left\{ {p}_{j} \in \mathcal{P} \mid \lVert {p}_{j} - {p}_{c} \rVert_{2} < r \right\}.
\end{equation}
The mean position $\bar{{p}}$ of these local points is computed as:
\begin{equation}
    \bar{{p}} = \frac{1}{N} \sum_{{p}_{j} \in \mathcal{N}} {p}_{j}.
\end{equation}
Then, all points are normalized by the mean value $\bar{{p}}$ and denoted as
\begin{equation}
\mathbf{X} = [p_1 - \bar{p}, p_2 - \bar{p}, \ldots, p_N - \bar{p}]^\top_{N \times 3}.
\end{equation}
Performing SVD on the centered data matrix $\mathbf{X}$
\begin{equation}
\mathbf{X} = \mathbf{U} \Sigma \mathbf{V}^\top
\end{equation}
where
$\mathbf{V} = [\mathbf{v}_1, \mathbf{v}_2, \mathbf{v}_3] \in \mathbb{R}^{3 \times 3}$, $\Sigma = [\text{diag}(\sigma_1, \sigma_2, \sigma_3), \mathbf{0}] \in \mathbb{R}^{N \times 3}$, $\sigma_1 \geq \sigma_2 \geq \sigma_3 $.
The right singular vector $\mathbf{v_1}$, corresponding to the principal direction of maximum variance in the local point cloud neighborhood $\mathcal{N}$, defines the optimal local grasp orientation. 
This orientation, aligned with the axial direction of the local rope segment, maximizes the contact area between the gripper and the rope, thereby ensuring a stable grasp.

\begin{figure*}[htbp]
    \centering
    \includegraphics[width=1\linewidth]{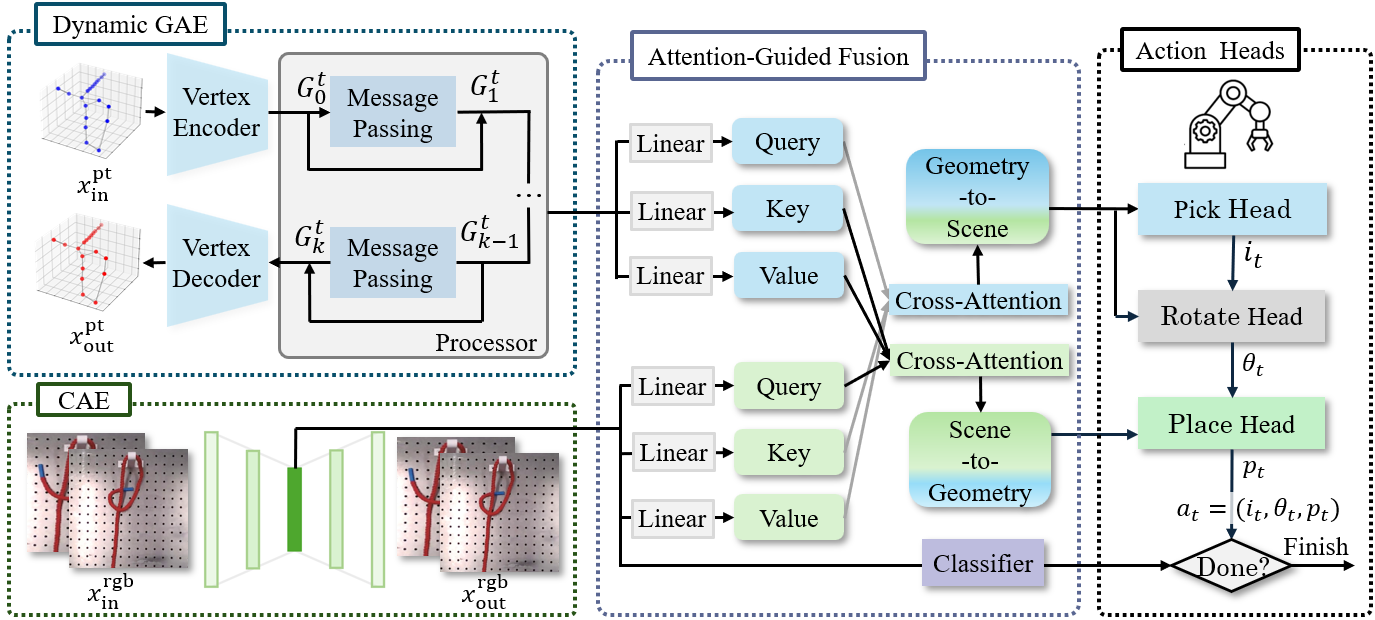}
    \caption
    {Overview of the proposed hitch knot tying framework. The system takes rope keypoints and a RGB image as input. A pretrained permutation-equivariant GAE processes keypoints to capture the geometry features, while a pretrained CAE extracts visual features. A bidirectional cross-attention module fuses these multimodal features. Three specialized heads then hierarchically predict the action tuple $a_t = (i_t, \theta_t, p_t)$: the Pick Head selects the grasp point, the Rotation Head predicts in-hand rotation, and the Place Head generates a placement location.
    }
    \label{framework}
\end{figure*}

\section{Hitch Knots Tying Framework}
\label{sec3_3}
The Hitch Knots Tying Framework, illustrated in Fig.~\ref{framework} and Algorithm~\ref{algorithm}, consists of three specialized modules: (1) a Multimodal Feature Extraction module employs a dynamic GAE to extract geometry features from $x^\text{pt}$ and a CAE to extract scene features from $x^\text{rgb}$;
(2) an Attention-Guided Fusion module that facilitates cross-modal reasoning through bidirectional cross-attention; and (3) Pick and Place Affordance Heads that decode the fused features into action parameters $(i_t, \theta_t, p_t)$.
This structured approach enables effective policy learning by leveraging both geometric and visual information with limited data.

\subsection{Multimodal Feature Extraction}

We self-supervise the extraction of rope keypoint (PT) and RGB image features in the knotting scenarios via reconstruction errors from the GAE and CAE. 
Given paired samples $\{x^\text{pt}, x^\text{rgb}\}$, the source features are represented as $\{f^{\text{pt}} \in \mathbb{R}^{n_p \times c^{pt}}, f^{\text{rgb}} \in \mathbb{R}^{h \times w \times c^{rgb}}\}$.

\subsubsection{Dynamic Graph Autoencoder} The GNN performs message passing between vertices and edges, making it particularly suitable for encoding the topological features of DLOs than general neural network \cite{wang2022offline}.
To address the \textit{Point Set Permutation} problem and achieve consistent rope feature representation, we employ a permutation-equivariant dynamic GAE. The symmetric message passing mechanism guarantees consistent feature encoding regardless of input keypoint ordering.
Specifically, we construct a dynamic graph $\mathcal{G}_t = (\mathcal{V}_t, \mathcal{E}_t)$ at each timestep, where vertices $\mathcal{V}_t$ correspond to the keypoints $x^\text{pt}$ and edges $\mathcal{E}_t$ are dynamically created between vertices within a Euclidean distance threshold. This adaptability allows the graph topology to effectively handle occlusions and deformations.
The vertex update rule in our GAE is designed to be permutation-invariant
\begin{equation}
 x_i' = \text{AGG}_{j \in \mathcal{N}(i)} \, \big(\text{MLP}\big(x_i \parallel \left(x_j - x_i\right) \big) \big)
\end{equation}
where $\mathcal{N}(i)$ is the set of neighbors of vertex $i$, $x_i$ denotes current keypoint position, and $x_j - x_i$ represents the relative position vector.  The symbol \(\parallel\) denotes concatenation. All edges share identical learnable parameters through a common multilayer perceptron (MLP) and symmetric aggregation functions (AGG), ensuring that the learned features depend solely on the underlying geometry rather than the arbitrary ordering of the keypoints \(x^\text{pt}\).
The GAE is trained with a reconstruction loss to learn a latent encoding $f^\text{pt}$ that effectively captures the rope's topological state.

\subsubsection{Image Autoencoder} To extract crucial features from high-dimensional visual information related to multiple actions, we utilize a CAE, which is adept at processing raw images. The CAE consists of an encoder \(\phi^I\) formed by convolutional and fully connected layers, along with a decoder \(\psi^I\) composed of deconvolutional and fully connected layers. During training, the parameters are optimized to  minimize the reconstruction error 
\begin{equation}
\phi^{I*}, \psi^{I*} = \arg \min_{\phi^I, \psi^I} \| x^\text{rgb} - (\phi^I \circ \psi^I)(x^\text{rgb}) \|^2,
\label{eq:recon}
\end{equation}
which allows the network to learn an optimal latent space \(f^\text{rgb} = \phi^I(x^\text{rgb})\) that captures essential information from original image. These image latent features are simultaneously employed to predict task termination through a lightweight MLP-based classifier.

\subsection{Attention-Guided Fusion}

While the permutation invariance of our dynamic GAE ensures consistent output features regardless of the observed keypoint ordering, the resulting representation still lacks an inherent physical information along the rope continuum. To address this \textit{Topological Ambiguity}, we integrate PT features with RGB features, which provide complementary visual context regarding the rope's global configuration and spatial relationships.

However, directly concatenating geometric PT features with visual RGB features can lead to spatial-semantic misalignment due to their heterogeneous representations. To enhance the reasoning capabilities from both geometric and visual representations, we employ a bidirectional cross-attention mechanism. 
This architecture dynamically associates keypoints with relevant image regions, reducing reliance on an explicit ordered keypoint sequence which is often fragile under occlusion or tracking failures.
In the geometry-to-scene pathway \(f^{\text{pt} \rightarrow \text{rgb}}\), queries derived from keypoint features attend to image features \(f^{\text{rgb}}\) (keys and values), thereby enriching each keypoint with visual context essential for predicting \textit{where to pick}. Conversely, in the scene-to-geometry pathway \(f^{\text{rgb} \rightarrow \text{pt}}\), queries derived from image features attend to keypoint features \(f^{\text{pt}}\) (keys and values), incorporating each image pixel with rope geometric information, crucial for predicting \textit{where to place}.

This cross-modal fusion ensures that both geometric and visual information are synergistically integrated, enabling the model to resolve topological ambiguities through complementary cues while maintaining a spatially consistent understanding of the rope's configuration.



\subsection{Action Affordance Head}

The action is parameterized as a tuple \(a_t = (i_t, \theta_t, p_t)\), where \(i_t\) denotes the index of the selected rope keypoint to grasp, \(p_t\) specifies the pixel-level placement location, and \(\theta_t\) determines the rotation on the z-axis from the pick point to the place point. This action space allows for the execution of complex knotting processes, such as looping, passing, and tightening, using a single-arm robot with a parallel gripper, eliminating the need for handovers or bimanual coordination.

To facilitate learning, each action dimension is discretized. The pick point is chosen from \(N_k\) perceived keypoints, the rotation is quantized into \(N_r\) bins covering a range from \(-90^\circ\) to \(90^\circ\), and the placement location is discretized at the image pixel level. Instead of directly regressing continuous action coordinates for the grasp and placement points, our approach computes a pick affordance over the rope keypoints and a pixel-wise affordance heatmap for placement. This formulation treats action prediction as a classification problem, which simplifies training \cite{combing2017} and inherently accommodates multimodality in the action space, arising from multiple actions that can transition the rope from the same initial to final configuration.

For action prediction, a naive approach might independently classify each action component. However, the pick-rotate-place operation exhibits strong conditional dependencies. Specifically, the choice of pick point influences feasible rotations, which subsequently constrain valid placement locations. To capture these dependencies without incurring the combinatorial complexity of the joint action space, we factor the action distribution as 
\begin{equation}
P(i_t, \theta_t, p_t) = P(i_t) P(\theta_t \mid i_t) P(p_t \mid i_t, \theta_t).
\end{equation}
This hierarchical decomposition is implemented through specialized network heads. The Pick Head processes the geometry-to-scene features \(f^{\text{pt} \to \text{rgb}}\) using a MLP followed by a softmax layer to generate a pick probability distribution across all keypoints. The selected pick point is determined as \(i_t = \arg\max_{i_t} P(i_t)\).
Subsequently, the Rotation Head utilizes the features of the chosen keypoint alongside \(f^{\text{pt} \to \text{rgb}}\) to predict the rotation angle \(\theta_t\). Finally, the Place Head combines one-hot encodings of the selected pick point and rotation with the vision-to-geometry features \(f^{\text{rgb} \to \text{pt}}\) and processes them through a decoder network to generate a spatial affordance heatmap \(\mathcal{H} \in \mathbb{R}^{H \times W}\). The placement location is then selected as the pixel with maximum affordance: \(p_t = \arg\max_{p_t} \mathcal{H}(p_t)\).

This cascaded design leverages geometrically enhanced features for picking and visually augmented features for placing. By explicitly modeling action dependencies, our approach simplifies learning and enhances policy stability.

\begin{algorithm}[t]
  \SetAlgoLined
  \KwIn{
  
  RGB Image: $x^\text{rgb}$, 
  
  Depth Image: $x^\text{depth}$, 
  
  Task Finish Flag: $done$\;} 
  \KwOut{
  
  Robot Action $a_t$;} 
  \textbf{Initialization}:\ 
  $done\xrightarrow{}False$\;
  Initialize $M_\text{PT}, M_\text{RGB}, M_\text{CLS}$ with pretrained weights;\
  
    \While{$not \; done $}{
    \tcc{---- Observation Phase -----}
    $x^\text{PT} = RopeDectction(x^\text{rgb}, x^\text{depth})$\;
    $x^\text{pt} = RopeNodeClustering(x^\text{PT})$\;
    \tcc{----  Manipulation Phase  ----}
    $ f^\text{pt}, f^\text{rgb} = M_\text{PT}(x^\text{pt}), M_\text{RGB}(x^\text{rgb})$\;
    $ f^{\text{pt}\to \text{rgb}}_{\text{cro}} = M_{FUSE}(f^\text{pt}, f^\text{rgb})$\;
    $ f^{\text{rgb}\to \text{pt}}_{\text{cro}} = M_{FUSE}(f^\text{rgb}, f^\text{pt})$\;
    $ i_t = PickHead( f^{\text{pt}\to \text{rgb}}_{\text{cro}} )$\;
    $ \overrightarrow{i_t} = SVD(i_t, x^\text{pt})$\;
    $ \theta_t = RotateHead(i_t, f^{\text{pt}\to \text{rgb}}_{\text{cro}})$\;
    $ p_t = PlaceHead(i_t, \theta_t, f^{\text{rgb}\to \text{pt}}_{\text{cro}})$\;
    $ a_t = (\overrightarrow{i_t}, \theta_t, p_t)$\;
    $ RobotCartesianMove(a_t)$\;
    $ done = M_\text{CLS}(f^\text{rgb})$\;
    }
  \caption{Algorithm For Hitch Rope Tying}
  \label{algorithm}
\end{algorithm}

\begin{table}[htbp]
\centering
\caption{PARAMETER CONFIGURATIONS OF THE PROPOSED APPROACH}
\label{tab:parameters}
\begin{tabular}{|l|l|}
\hline
\multicolumn{2}{|c|}{\textbf{GAE Model Parameters}} \\
\hline
\textit{Encoder} & \\
\quad Hidden Layers & 1 \\
\quad Hidden Size & 64 \\
\quad Activation Function & ReLU \\
\textit{Processor} & \\
\quad Graph Construction & k-NN (k=3) \\
\quad Message Passing Steps ($k$) & 2 \\
\quad Hidden Layers & 2 \\
\quad Hidden Size & 64 \\
\quad Activation Function & ReLU \\
\textit{Decoder} & \\
\quad Hidden Layers & 1 \\
\quad Hidden Size & 64 \\
\quad Activation Function & ReLU \\
\hline
\multicolumn{2}{|c|}{\textbf{CAE Model Parameters}} \\
\hline
\textit{Encoder} & \\
\quad Convolution Layers & 3 \\
\quad Channels & $32 \rightarrow 64 \rightarrow 128$ \\
\quad Kernel Size & $3 \times 3$ \\
\textit{Decoder} & \\
\quad Transpose Conv Layers & 3 \\
\quad Channels & $64 \rightarrow 32 \rightarrow 3$ \\
\quad Kernel Size & $3 \times 3$ \\
\hline
\multicolumn{2}{|c|}{\textbf{Classifier \& Data Fusion}} \\
\hline
\textit{Classifier} & \\
\quad Hidden Layers & 2 \\
\quad Hidden Size & 64 \\
\textit{Data Fusion} & \\
\quad Query/Key/Value Dimensions & 64 \\
\hline
\multicolumn{2}{|c|}{\textbf{Data \& Training Parameters}} \\
\hline
Batch Size & 8 \\
Image Resolution & $640 \times 480$ \\
Keypoint Number ($N_k$) & 20 \\
$\lambda_\text{pick}$, $\lambda_\text{rot}$, $\lambda_\text{place}$ & 2, 1, 2 \\
Learning Rate & $5 \times 10^{-4}$ \\
Optimizer & Adam \\
Train/Test Split Ratio & 0.8 \\
\hline
\end{tabular}
\end{table}

\section{Experiment and Results}
\label{sec4}
\begin{figure*}[t]
    \centering
    \includegraphics[width=1.0\linewidth]{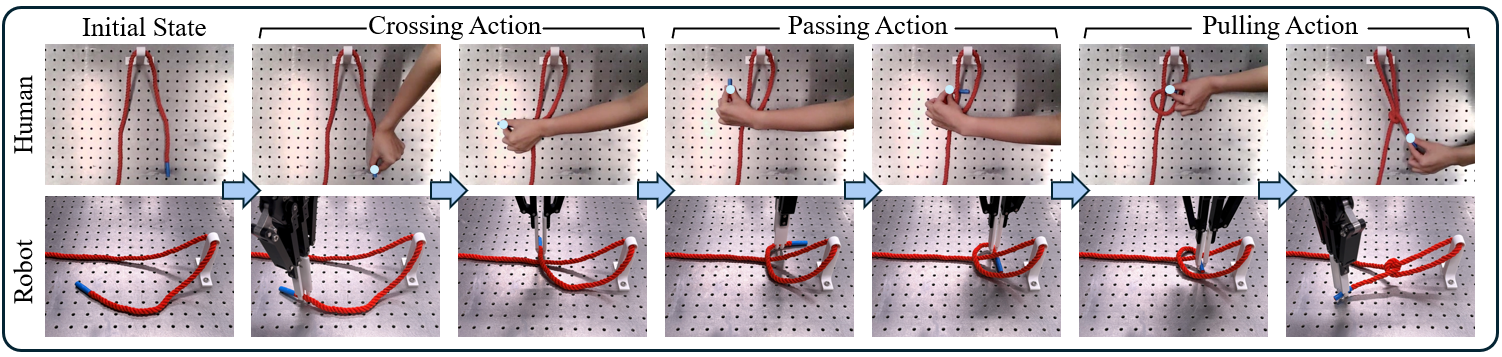}
    \caption{ Examples of sequential motions in hitch knot tying using nylon rope. The human hand motions (top) are derived from the trajectories of the thumb and index finger tips, captured with MediaPipe. The robot (bottom) dynamically infers pick and place actions based on perceptual feedback. 
    }
    \label{example}
\end{figure*}

\subsection{Hardware Setting}
\label{sec4_1}

\begin{figure}[htbp]
    \centering
    \includegraphics[width=1.0\linewidth]{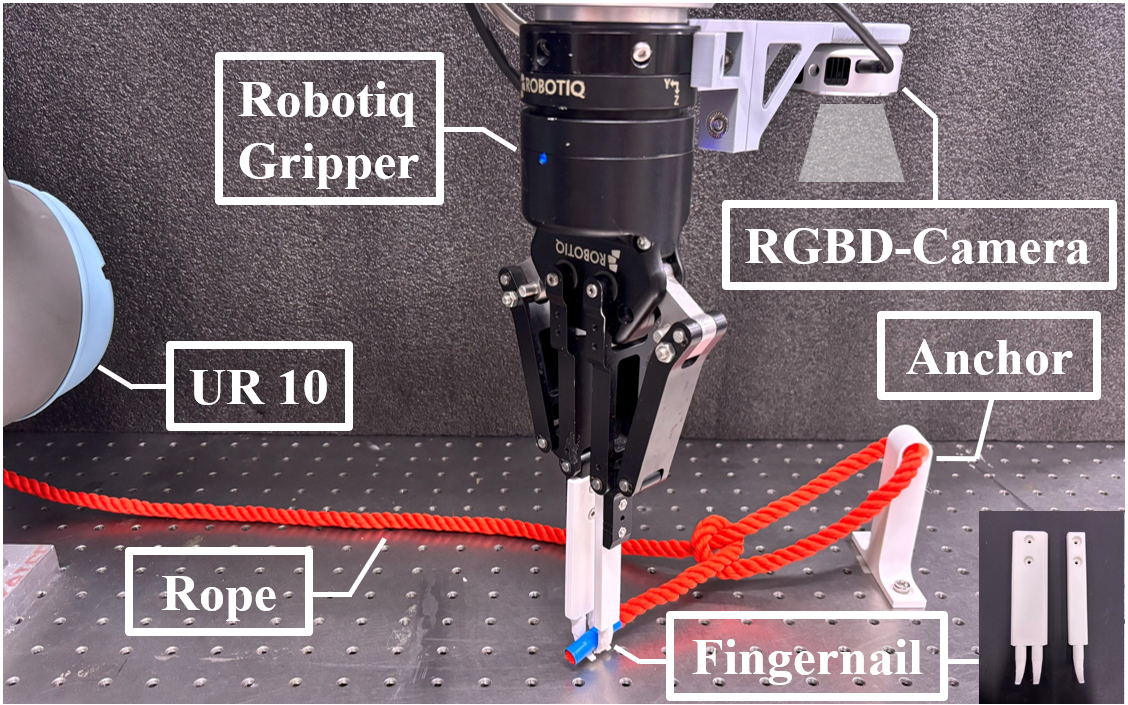}
    \caption{Hardware setup for conducting rope tying experiments. }
    \label{hardware}
\end{figure}

Fig. \ref{hardware} shows our experimental setup, which consists of a 6-DOF UR 10 manipulator, a wrist-mounted Intel RealSense L515 RGBD camera, and a Robotiq 2F-140 gripper equipped with bio-inspired fingernails\cite{Zuo2025} to enhance grasping performance. 
The rope can be initialized in either a crossed or uncrossed shape, and it is distinguished from the background using depth and color filtering.
One tip of the rope is marked with blue tape, only to indicate the working end during human manipulation.

\subsection{Training via Demonstration}
We collected 100 human demonstration videos and utilized MediaPipe\cite{zhang2020mediapipe} to extract action points and rotation angles based on the positions and orientations of the thumb and index finger tips, as illustrated in the first row of Fig.~\ref{example}. From these demonstrations, we derived a set of expert trajectories \(\{\tau_1, \tau_2, \ldots\}\), where each trajectory \(\tau\) consists of a sequence of demonstrated state-action pairs \(\{(s_0, a_0), (s_1, a_1), \ldots, (s_n, a_n)\}\), forming our training dataset.
In this dataset, the pick affordance of a rope is represented as a one-hot vector, with the element corresponding to the node nearest to the grasping point set to 1. Conversely, the place affordance map is modeled as a two-dimensional Gaussian probability distribution centered at the target placement point. This dataset was then utilized to train the network using a behavior cloning algorithm \cite{argall2009survey}, with the training process conducted on a computer equipped with an NVIDIA GeForce RTX 4080 GPU.

The total loss for the framework is defined as a weighted sum of the individual losses
\begin{equation}
\mathcal{L} = \lambda_\text{pick}\mathcal{L}_\text{pick} + \lambda_\text{rot}\mathcal{L}_\text{rot} + \lambda_\text{place}\mathcal{L}_\text{place},
\end{equation}
where \(\mathcal{L}_\text{pick}\) and \(\mathcal{L}_\text{rot}\) are cross-entropy losses, \(\mathcal{L}_\text{place}\) is a negative log-likelihood loss over the Gaussian-smoothed place heatmap, and \(\lambda\) represents the weighting factors for the different losses. The parameter configurations of the proposed method are detailed in Table \ref{tab:parameters}.

\subsection{Performance Evaluation}
\label{sec4_2}
We evaluate our method through a series of real-world experiments, achieving an overall success rate of 84\% (42/50 trials) for hitch knot tying with the trained rope. This performance surpasses existing methods as compared in Table~\ref{tab:comparison}. Notably, our approach outperforms both previous model-based and learning-based techniques in comparable single-arm settings. Fig.~\ref{example} depicts sequential examples of the knotting process using a nylon rope, highlighting the robot's capability to dynamically select pick points from disordered keypoints and infer feasible placement locations toward the target configuration. 

To further assess the generalizability of our method, we varied the rope material, diameter, and background settings, as outlined in Table~\ref{tab:generalization}. 
The initial state of the rope was always randomized. 
Our method generalizes well to unseen backgrounds (Scenario 2, 4/5), but performance declines with larger diameter (Scenario 3, 3/5) and fails entirely with polypropylene rope (Scenario 4, 0/5).
The failures are primarily attributed to high rope stiffness, resulting from both increased diameter and material differences. Polypropylene exhibits significantly higher stiffness than nylon, leading to strong elastic restoration during bending. This restoration prevents the rope from maintaining the deformed configuration required for knotting, often resulting in task termination. These results indicate a limitation in handling highly stiff materials and emphasize the need for dual-arm coordination.

\begin{table}[htbp]
\caption{COMPARISON WITH OTHER KNOTTING METHODS}
\label{tab:comparison}
\centering
\begin{tabular}{ccccc}
\toprule
Reference & Method &  Arm & Success Rate  \\
\midrule
Dinkel \textit{et al.} \cite{dinkel2025knotdlo} &   Model & Single & 50\% \\
Nair \textit{et al.} \cite{combing2017} & Learning & Single & 38\% \\
Priya \textit{et al.} \cite{9197121} & Learning & Single & 66\% \\
Pathak \textit{et al.} \cite{8575448} & Learning & Single & 60\% \\
\textbf{RopeHitch(Ours)} & Learning & Single & \textbf{84\%}\\
\bottomrule
\end{tabular}
\end{table}

\begin{table*}[htbp]
\caption{Generalization Evaluation with Our Knotting Methods}
\label{tab:generalization}
\centering
\begin{tabular}{ ccccc}
\toprule
&  Scenario 1  &  Scenario 2  &  Scenario 3  &  Scenario 4  \\
\midrule
 Scenarios  &
\includegraphics[width=3cm, height=2cm]{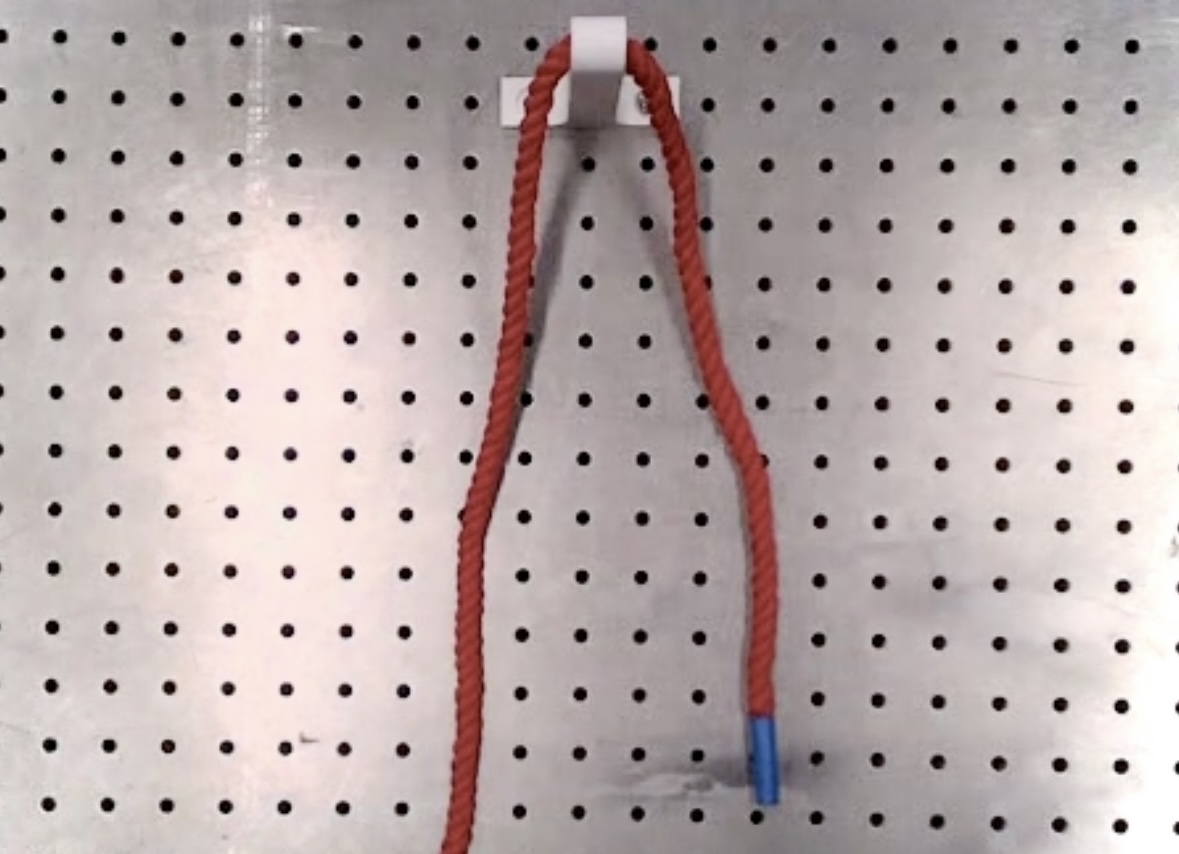} &
\includegraphics[width=3cm, height=2cm]{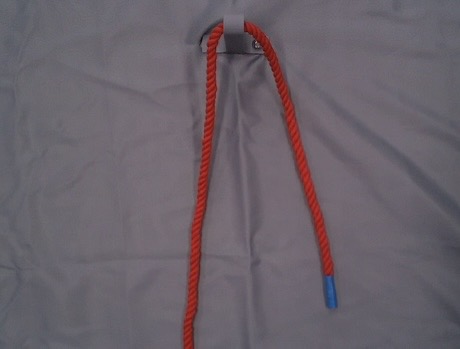} &
\includegraphics[width=3cm, height=2cm]{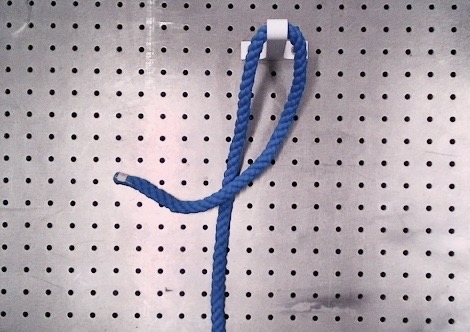} & 
\includegraphics[width=3cm, height=2cm]{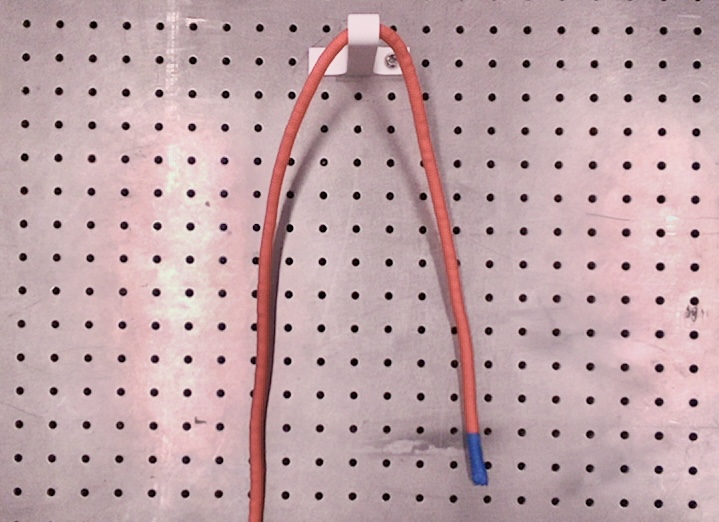} \\
 Diameter  & 10mm & 10mm & 14mm & 10mm \\
 Material  & Nylon & Nylon & Nylon & Polypropylene  \\
 Background  & seen & unseen & seen & seen \\
 Success Rate  & 5/5 & 4/5 & 3/5 & 0/5  \\
\bottomrule
\end{tabular}
\end{table*}

\subsection{Ablation Study}

This section presents ablation studies to evaluate the contribution of individual components within the proposed framework. We examine both PT and RGB modalities independently and compare our bidirectional cross-attention fusion strategy against a direct concatenation baseline.

\begin{figure}[htbp]
\centering
\includegraphics[width=1 \linewidth]{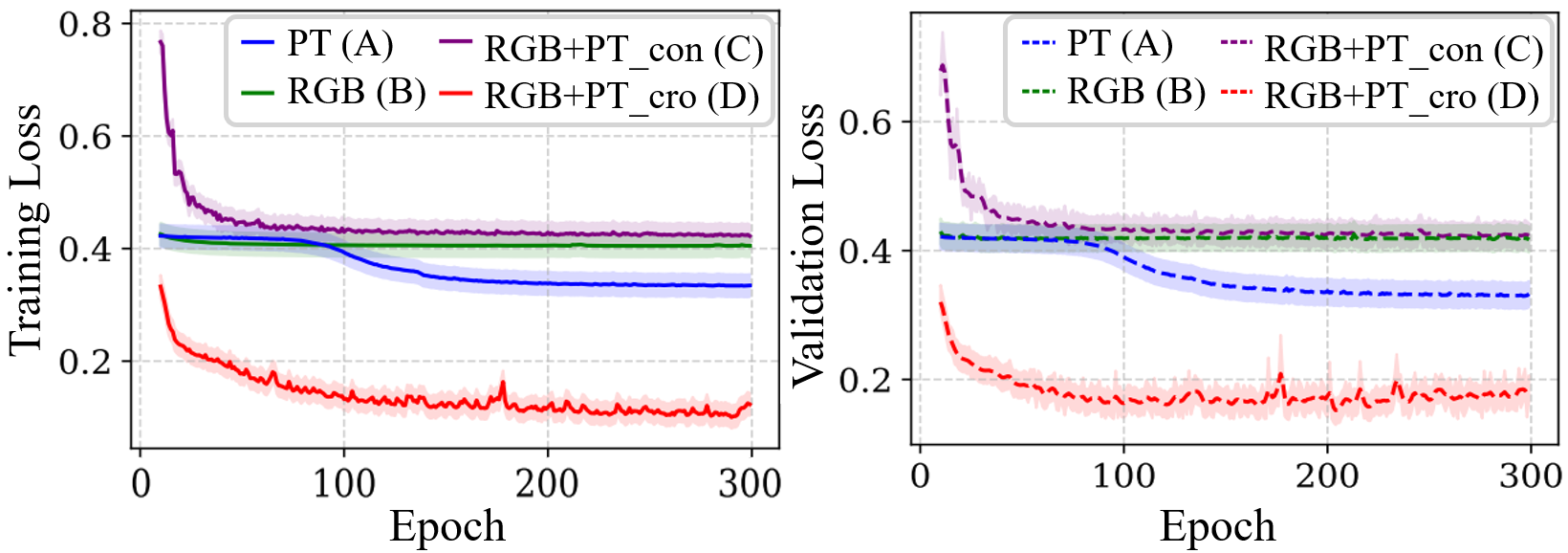}
\caption{Training and validation loss curves for different modality configurations. The PT-only input (case A) achieves lower loss than the RGB-only input (case B), while our attention-guided fusion approach (case D) outperforms direct concatenation method (case C).
}
\label{loss}
\end{figure}

\begin{figure}[htbp]
\centering
\includegraphics[width=1 \linewidth]{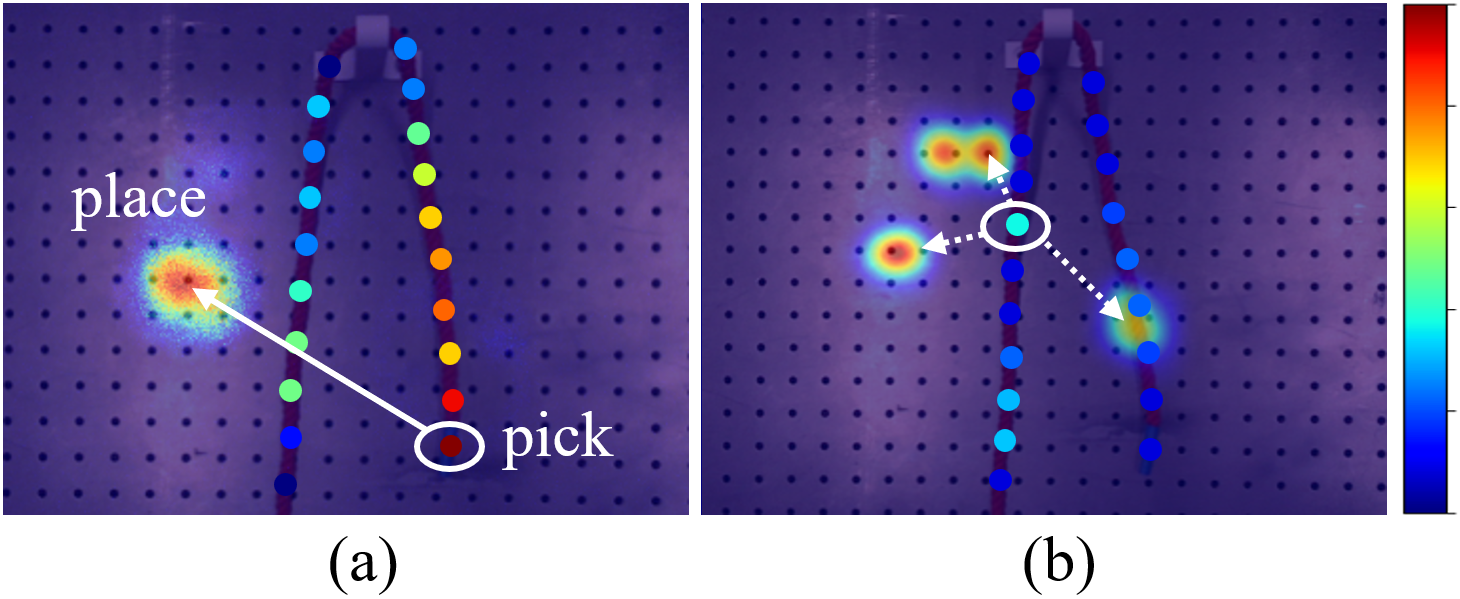}
\caption{
Comparative analysis of action affordance predictions. (a) Our fused RGB+PT modality accurately predicts kinematically feasible grasp points and well-localized placement regions. (b) In contrast, using RGB-only for grasping and PT-only for placement results in infeasible grasp points and dispersed placement affordances.
}
\label{visual}
\end{figure}

Fig.~\ref{loss} shows the training loss curves over 300 epochs. The PT-only modality (case A) achieves a lower training loss compared to the RGB-only modality (case B), indicating that geometric keypoints encode task-relevant features more effectively than visual data alone. Furthermore, our cross-attention fusion approach (case D) achieves a lower final loss than direct concatenation (case C), supporting our hypothesis that naive feature concatenation may lead to spatial-semantic misalignment.
Affordance visualization in Fig.~\ref{visual} reveals complementary characteristics of the modalities: PT-only inputs enable precise grasp affordances but yield ambiguous placement predictions, while RGB-only inputs provide clear placement cues but struggle with accurate grasp localization. Our fused representation (case D) effectively integrates geometric and visual cues, enabling robust predictions for both action steps.
As summarized in Table~\ref{tab:ablation_study}, our approach achieves successful outcomes, while other configurations failed completely. This demonstrates the critical importance of effective multimodal fusion.

\label{sec4_3}
\begin{table}[htbp]
\caption{ABLATION STUDY ON DIFFERENT MODULES}
\label{tab:ablation_study}
\centering
\small
\begin{threeparttable} 
\begin{tabular}{lcccc}
\toprule
\multirow{2}{*}{Module/Case} & \multicolumn{4}{c}{Case} \\
\cmidrule(lr){2-5}
 & A & B & C & D \\
\midrule
PT modality & $\checkmark$ & $\times$ & $\checkmark$ & $\checkmark$ \\
RGB modality & $\times$ & $\checkmark$ & $\checkmark$ & $\checkmark$ \\
Attention & $\times$ & $\times$ & $\times$ & $\checkmark$ \\
\midrule
Success/Fail & F & F & F & S \\
\bottomrule
\end{tabular}
\begin{tablenotes}
\footnotesize
\item {*PT and RGB indicate input modalities, while Attention denotes cross-attention.}
\end{tablenotes}
\end{threeparttable} 
\end{table}


\section{Conclusion}

In this paper, we propose a novel framework to construct relatively robust multimodal representations for hitch knot tying.  
Our approach involves clustering disordered rope keypoints from the rope point cloud and obtaining PT and RGB embeddings through self-supervised learning with a dynamic GAE and a CAE.  
The bidirectional cross-attention module effectively incorporates implicit interactions between the RGB and PT modalities, enabling these cross-modal embeddings to predict action affordances for manipulating the rope in various states. 
Generalization experiments have confirmed the effectiveness of our method in the hitch knot tying task, and ablation studies have been conducted to evaluate the contribution of individual components.
However, our method is still challenged by low-quality data and the limitations imposed by single-arm manipulation. 
Future work will focus on developing more robust perception and manipulation methods under complex conditions to enhance our framework's performance.

\bibliographystyle{ieeetr} 
\bibliography{reference}

\end{document}